\documentclass[final]{cvpr}

\usepackage{times}
\usepackage{epsfig}
\usepackage{graphicx}
\usepackage{amsmath}
\usepackage{amssymb}
\usepackage{multirow}
\usepackage{tabularx}
\usepackage{balance} 
\usepackage{bm}
\usepackage{comment}
\usepackage[pagebackref=true,breaklinks=true,colorlinks,bookmarks=false]{hyperref}

\def\cvprPaperID{4182} 
\def\confYear{CVPR 2021}

\begin{document}

\title{Self-supervised Video Retrieval Transformer Network}

\author{Xiangteng He, Yulin Pan, Mingqian Tang and Yiliang Lv\\
Alibaba DAMO Academy\\
}

\maketitle

\begin{abstract}
    Content-based video retrieval aims to find videos from a large video database that are similar to or even near-duplicate of a given query video. It plays an important role in many video related applications, including copyright protection, recommendation, ﬁltering and etc.. Video representation and similarity search algorithms are crucial to any video retrieval system. To derive effective video representation, most video retrieval systems require a large amount of manually annotated data for training, making it costly inefficient. In addition, most retrieval systems are based on frame-level features for video similarity searching, making it expensive both storage wise and search wise. We propose a novel video retrieval system, termed SVRTN, that effectively addresses the above shortcomings. It first applies self-supervised training to effectively learn video representation from unlabeled data to avoid the expensive cost of manual annotation. Then, it exploits transformer structure to aggregate frame-level features into clip-level to reduce both storage space and search complexity. It can learn the complementary and discriminative information from the interactions among clip frames, as well as acquire the frame permutation and missing invariant ability to  support more flexible retrieval manners. Comprehensive experiments on two challenging video retrieval datasets, namely FIVR-200K and SVD, verify the effectiveness of our proposed SVRTN method, which achieves the best performance of video retrieval on accuracy and efficiency.
\end{abstract}

\section{Introduction}
\begin{figure}[!t]
  \begin{center}\includegraphics[width=1\linewidth]{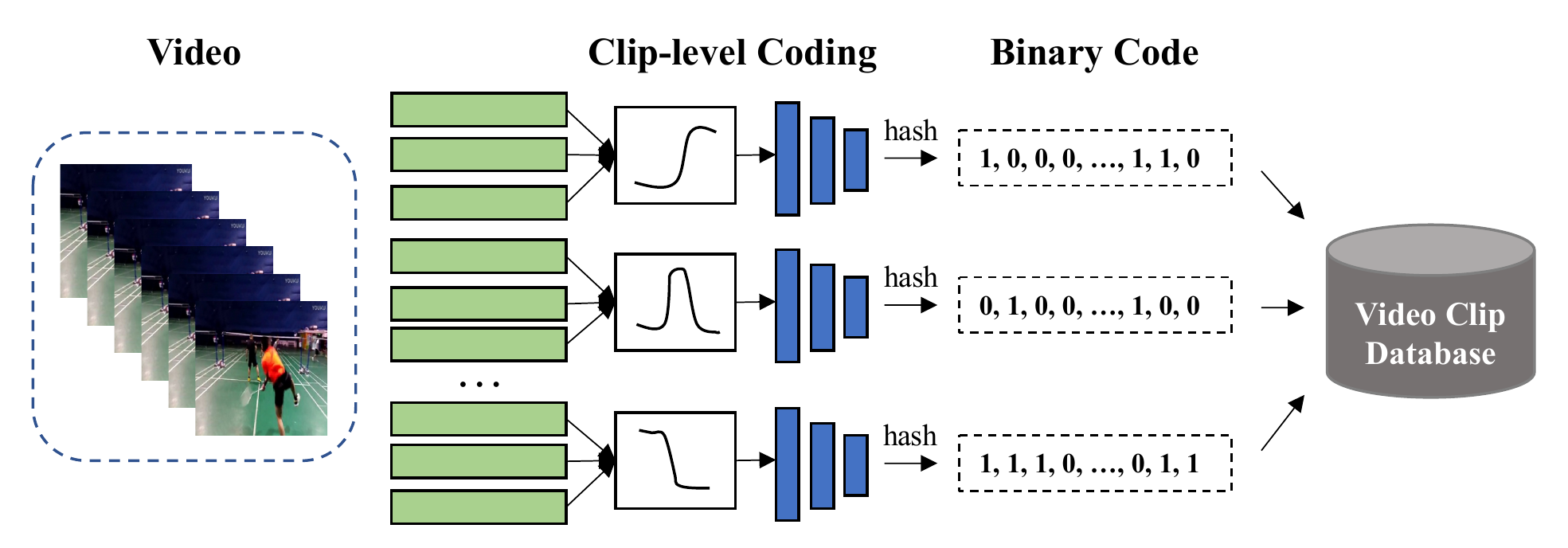}
  \caption{The illustration of clip-level video representation. Videos are split into shots via shot boundary detection, and further divided into clips at a fixed time interval. Then our SVRTN approach is applied to extract clip-level video representation. Finally, we binarize the clip-level features via hashing method for efficient retrieval.}
  \label{fig:clip-level}
  \end{center}
\end{figure}
These days, we have witnessed dramatic increase in the volume of videos generated over internet. At the same time, we have also observed a large number of videos that essentially steal contents from others, making video copyright protection and filtering an important demand. 
Content-based video retrieval addresses the problem by identifying a subset of videos from a large video database, which share similar contents with a given query video. It has drawn much attention, such as near-duplicate video retrieval \cite{jiang2019svd} and fine-grained incident video retrieval \cite{kordopatis2019fivr}. 

To design an efﬁcient and effective video retrieval system, two components are important, i.e. video representation and video search. For video representation, most existing methods \cite{kordopatis2017near,kordopatis2019visil} apply supervised deep learning technologies to learn appropriate feature representation for accurate video content matching. Since a large amount of labeled videos are needed for training, it is costly to learn a robust and powerful video representation in this way. 

For video search, most methods \cite{kordopatis2019visil,kordopatis2017near2,chou2015pattern,tan2009scalable} represent each video by a set of frame-level features, and the similarity between two videos are decided by the similarities between frames from the two videos followed by temporal alignment analysis, such as dynamic programming\cite{chou2015pattern,liu2017image}, temporal network\cite{tan2009scalable,jiang2016partial,hu2018learning}, and Temporal Hough Voting\cite{jiang2014vcdb,douze2010image}. The main shortcoming of these methods are of two folders. First, it needs to store all the frame-level features from all videos, making it storage expensive. Second, since the similarity measurement between two videos requires the similarity measurement between frames, making it computationally expensive. One common approach  \cite{kordopatis2017near,shao2020context} to address these limitations is to represent each video by a single vector (i.e. video-level features). Although these approaches help alleviate the problems of storage and computational cost, as pointed out in \cite{song2011multiple}, they are insufficient to capture crucial details of individual videos, particularly for long videos. 

To address the above problems, we propose a Self-supervised Video Retrieval Transformer Network (SVRTN). It leverages self-supervised learning to learn video representation from unlabeled data, and exploits transformer structure to aggregate frame-level features into clip-level, as shown in Figure \ref{fig:clip-level}. More specifically, the key contributions of this work can be summarized as follow;
\begin{itemize}
    \item 
    \textbf{Self-supervised video representation learning} is proposed to learn the representation with the pairs of the video and its transformations, which are automatically generated by temporal and spatial transformations, thus avoiding the high cost in manual annotation.  Due to the self-generation of training data, our SVRTN approach can learn better video representation from a large amount of unlabelled videos, leading to better generalization for its learned representation.
    \item
    \textbf{Clip-level set transformer network} is proposed to aggregate frame-level features into clip-level, leading to significant reduction in both storage space and  search complexity.  It can learns the complementary and variant information from the interactions among clip frames via self-attention mechanism, as well as acquires the frame permutation and missing invariant ability to handle the issue of missing frames, both of which increase the discrimination and robustness of the clip-level feature. Besides, it supports more flexible retrieval manners, such as clip-to-clip retrieval and frame-to-clip retrieval.
\end{itemize}

Comprehensive experiments on two challenging video retrieval datasets, namely FIVR-200K and SVD, verify the effectiveness of our SVRTN approach, which achieves the best performance of video retrieval on accuracy and efficiency. Compared with video-level methods, our SVRTN achieves the improvements of 30.6\%, 28.2\%, 21.3\% mAPs on the DSVR, CSVR and ISVR tasks of FIVR-200K dataset, and 4.7\% mAP on SVD dataset. Compared with frame-level methods, our SVRTN approach achieves comparable performance, and reduces about 78.7\% of the feature storage cost and increases the retrieval speed by $\sim 25$ times.

\section{Related Work}
Existing video retrieval methods can be divided into two categories: frame-level retrieval methods and video-level retrieval methods.

\subsection{Frame-level Retrieval Methods}
These methods generally extract frame-level features using CNN, and retrieve related frames by approximate nearest neighbor search. Various post-processing methods\cite{jiang2014vcdb,tan2009scalable,hu2018learning,feng2018video,baraldi2018lamv,revaud2013event} have been proposed to aggregate the frame-to-frame similarity matrix to video similarity score. Jiang et al. propose Temporal Hough Voting \cite{jiang2014vcdb} to find temporal alignments, which makes full use of the relative timestamp between matched frames. Tan et al. propose Graph-based Temporal Network \cite{tan2009scalable} to detect the longest shared path between two compared videos. Hu and Lu \cite{hu2018learning} combine temporal network with a CNN+RNN feature encoder, to address the problem of partial copied detection. Another popular solution is based on Dynamic Programming(DP), which is applied to extract the biggest matched diagonal block from frame-to-frame similarity matrix, and tolerate limited horizontal and vertical movements for flexibility.
Chou et al. \cite{chou2015pattern} apply Bag-of-Words to represent frames, and propose m-pattern-based dynamic programming (mPDP) algorithm to
localize near-duplicate segments and re-rank the retrieved videos. 
However, the above methods ignore exploiting spatial feature invariance, which is essential to video retrieval. Recently, Kordopatis-Zilos et al. \cite{kordopatis2019visil} employ a region-level similarity calculation and aggregate region similarity matrix to frame similarities, which considers fine-grained spatial alignments and achieves high retrieval performance. 
These frame-level retrieval methods disregard the redundancy between successive frames, so that more computation cost will be needed, resulting in a low retrieval efficiency. 

\begin{figure*}[!t]
  \begin{center}\includegraphics[width=0.89\linewidth]{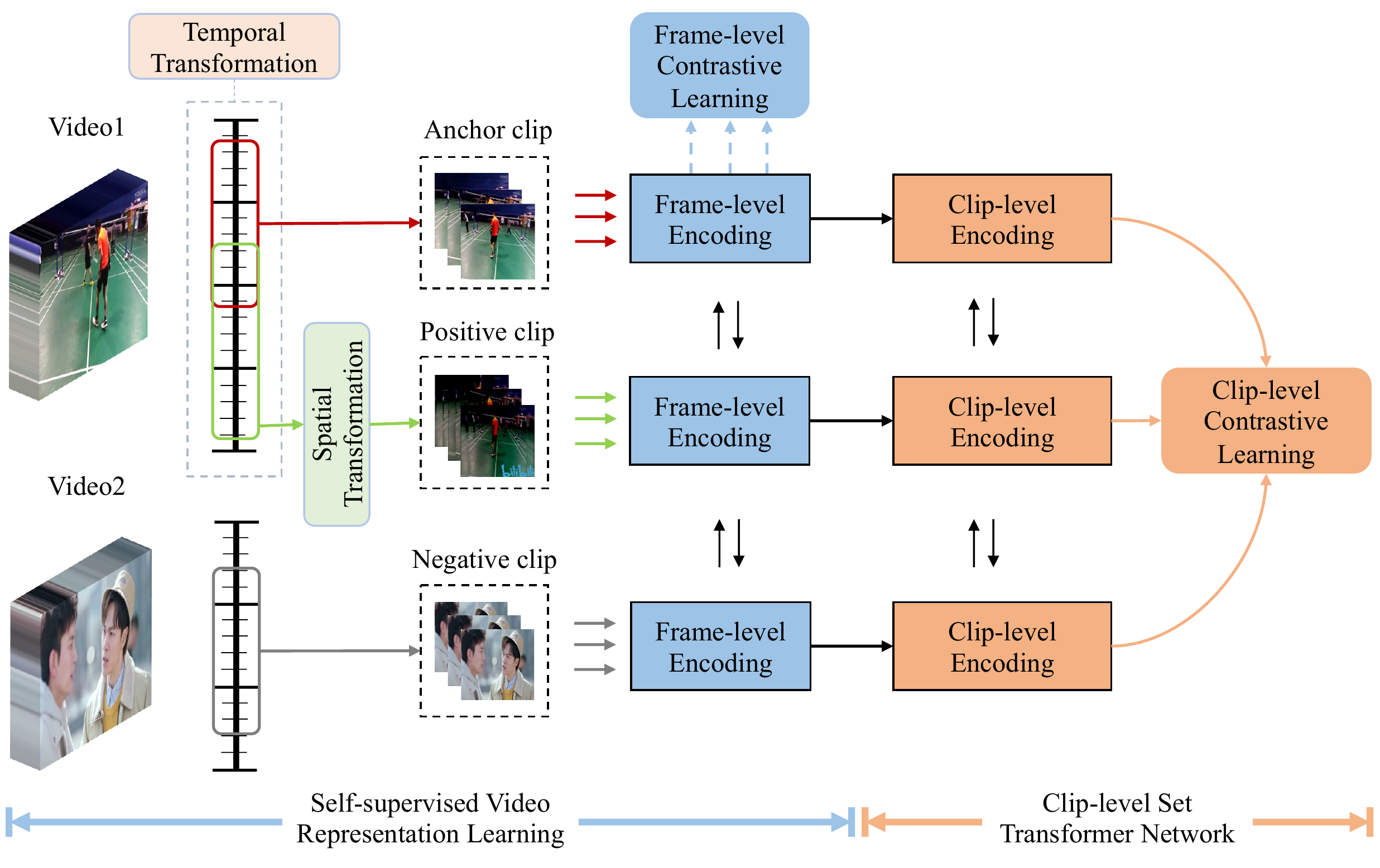}
  \caption{Overview of our SVRTN approach.}
  \label{fig:framework}
  \end{center}
\end{figure*}
\subsection{Video-level Retrieval Methods} 
Video-level retrieval methods encode the videos in video-level, and search for the $k$-nearest neighbors for the video-level feature of the query video in the embedding space. Various frame feature aggregation methods\cite{kordopatis2017near,kordopatis2017near2,liao2018ir,cai2011million,gao2017er3,wu2007practical} have been used  to obtain a single video-level representation. Liong et al. \cite{liong2016deep} propose temporal pooling layer to aggregate the successive frames by the means of average pooling. Kordopatis-Zilos et al. \cite{kordopatis2017near2} extract individual frame features from intermediate CNN layers, and adopt Bag-of-Words to compress them into a video-level  representation, so that video similarity can be measured by calculating the cosine distance between the two video-level representations. Furthermore, Kordopatis-Zilos et al. \cite{kordopatis2017near} aggregate frame features by the means of average pooling, and introduce Deep Metric Learning(DML) to learn an embedding by minimizing the distance between related videos and maximizing the distance between irrelevant ones. Hash codes based methods \cite{cai2011million,li2019neighborhood,hao2016stochastic,song2018self} are also widely used to encode unified spatial-temporal representation from videos. Song et al. \cite{song2018self} capture the temporal relationship between frames using an encoder-decoder architecture. Li et al. \cite{li2019neighborhood} apply the binary codes to capture spatial-temporal structure in a video by integrating the neighborhood attention mechanism into an RNN-based reconstruction scheme. However, these video-level retrieval methods generally perform worse than frame-level retrieval methods, which is mainly due to that single vector is hard to capture the entire spatio-temporal structure in a video sufficiently.

\section{Self-supervised Video Retrieval Transformer Network}
In this section, we present the proposed self-supervised video retrieval transformer network (SVRTN) for efficient retrieval by reducing the expensive cost of manual annotation, storage space and similarity search. It mainly consists of two components: self-supervised video representation learning and clip-level set transformer network, as shown in Figure \ref{fig:framework}. First, we automatically generate the video pairs via temporal and spatial transformations. Then, we utilize these video pairs as supervision to learn frame-level feature with contrastive learning. Finally, we aggregate the frame-level features into clip-level feature via self-attention mechanism, and increase the robustness via masked frame modeling.

\subsection{Self-supervised Video Representation Learning}
\begin{figure*}[!ht]
  \begin{center}\includegraphics[width=0.95\linewidth]{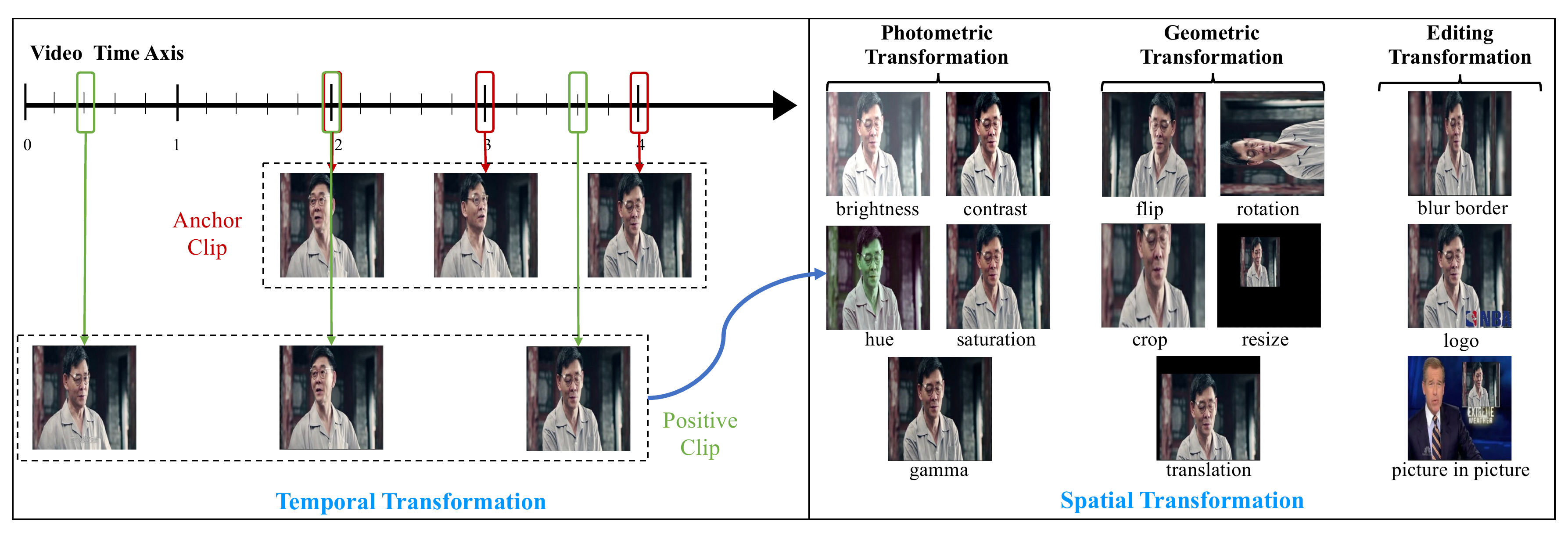}
  \caption{Illustration of self-generation of training data.}
  \label{fig:transformation}
  \end{center}
\end{figure*}
Existing methods generally train their model with manual annotated video pairs. The more data, the better performance will be achieved \cite{sun2017revisiting}. However, the cost of annotation is too expensive to generate a large amount of training data. So it is normal that representation learning is restricted to the limited volume of training data. Inspired by the advance of recent self-supervised learning methods \cite{gidaris2018unsupervised,kolesnikov2019revisiting,dosovitskiy2014discriminative}, we propose the self-supervised video representation learning to break the restriction, and exploit the spatial-temporal invariant of representation to defense various video transformations. 

\subsubsection{Self-generation of Training Data}
First, we automatically collect a large amounts of videos from the video website. Then, temporal and spatial transformations are sequentially performed on these clips to construct the training data.

\textbf{(1) Temporal Transformation}: 
As shown in the left part of Figure \ref{fig:transformation}, given a video, we first uniformly sample $N$ frames with a fixed time interval $r$ to generate the anchor clip, denoted as $C=\{{I}^{1},{I}^{2},\cdots,{I}^{N} \}$. Then a frame ${I}^{m}$ is randomly selected from the anchor clip as the identical content shared by anchor clip $C$ and positive clip $C_{+}$. We regard the selected frame as the median frame of $C_{+}$, and uniformly sample $\frac{N-1}{2}$ frames forward and backward respectively, with a different sample time interval $r_{+}$. 

\textbf{(2) Spatial Transformation}:
For each frame, we further perform spatial transformation. As shown in the right part of Figure \ref{fig:transformation}, three types of spatial transformations are considered: \emph{(a) Photometric transformation}. It includes the transformations of brightness, contrast, hue, saturation and gamma adjustment. \emph{(b) Geometric transformation}. It includes the transformations of horizontal flip, rotation, crop, resize and translation. \emph{(c) Editing transformation}. It includes the transformations of adding blurred background, adding logo, picture in picture and etc.. In training stage, we randomly select one transformation from each type of spatial transformation, and then apply them on frames from positive clips in sequence to generate the new positive clips.
\begin{figure}[!t]
  \begin{center}\includegraphics[width=1\linewidth]{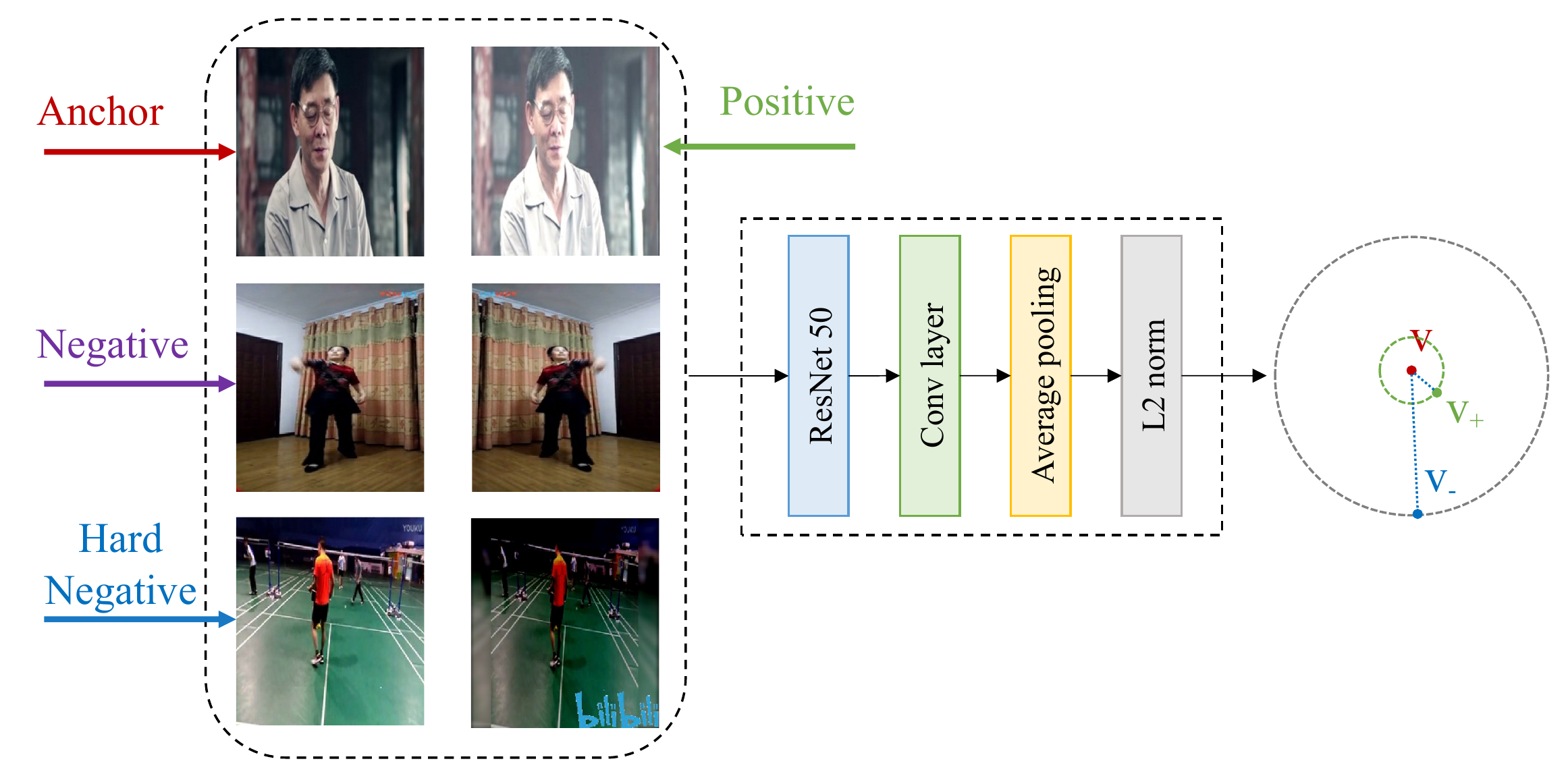}
  \caption{Model architecture.}
  \label{fig:backbone}
  \end{center}
\end{figure}

\subsubsection{Video Representation Learning}
\label{sub:videorepresentationlearning}
Since the supervised video pairs are generated, we use them to learn the video representation with frame-level contrastive loss. As shown in Figure \ref{fig:backbone}, we adopts ResNet50 \cite{he2016deep} as feature encoder, and then followed by a convolutional layer to reduce the channel number of the feature map, finally average pooling and $L_2$ normalization are applied to obtain the frame-level feature.

The goal of video representation learning is to capture spatial structure from individual frames and ignore the impacts of various transformations, through minimizing the distance between features of the anchor clip frames and positive clip frames, as well as maximizing the distance between features of the anchor/positive clip frames and negative clip frames.

Specifically, given an anchor clip containing $N$ frames $C = \{ {I}^{1},{I}^{2},\cdots,{I}^{N} \}$, then a positive clip is generated via temporal and spatial transformations, denoted as $C_+ =\{  {I}_{+}^{1},{I}_{+}^{2},\cdots,{I}_{+}^{N}\}$.
We organize these frames in semantic-related pairs $\{({I}^{t},{I}_{+}^{t}) \}_{t=1}^{N}$. Then video representation learning is employed to encode spatial structure from individual frames, which is formulated as 
\begin{equation}
{v} = f_{S}({I})
\label{eqa: sse_enc}
\end{equation}

Since a set of frame-level features $S_F=\{({v}^{t},{v}_{+}^{t})\}_{t=1}^{N}$ is obtained, a contrastive learning is adopted to drive the features more discriminative and robust. The loss function is an adapted noise contrastive estimation loss \cite{oord2018representation}, and its definition is as follows:
\begin{equation}
\begin{aligned}
L_{F} = \frac{1}{N} \sum_{t=1}^{N} &-\mathbb{E}_{P_{d}} \log P(D=1|{v}^{t}, {v}_{+}^{t}) \\
        &-(1-\mathbb{E}_{P_{d}}) \log(1-P(D=1|{v}^{t}, {v}_{+}^{t}))
\label{eqa:sc}
\end{aligned}
\end{equation}
where $P_{d}$ denotes the actual data distribution and $\mathbb{E}_{P_{d}}=1$ indicates ${I}^{t}$ and ${I}_{+}^{t}$ share absolutely identical visual semantic. The probability of the encoded vectors ${v}^{t}$ with ${v}_{+}^{t}$ is from the data distribution $P(D=1|{v}^{t}, {v}_{+}^{t})$ can be defined as :
\begin{equation}
P(D=1|{v}^{t}, {v}_{+}^{t}) = \frac{\exp({{v}^{t}}^\mathrm{T}{v}_{+}^{t})}{\exp({{v}^{t}}^\mathrm{T}{v}_{+}^{t})+\max\limits_{{v}_{-} \notin S_F}\exp({{v}^{t}}^\mathrm{T}{v}_{-})}
\label{eq:pd}
\end{equation}
where ${v}_{-}$ indicates the feature of frame from the negative chip, which is semantic-irrelevant with anchor clip. It is noted that only the batch-hardest negative frame will contribute to the $P(D=1|{v}^{t}, {v}_{+}^{t})$, because the simple negative frames will decrease the discriminability of the learned feature.

\subsection{Clip-level Set Transformer Network}
Since the adjacent frames from one clip have the similar content, the frame-level features have high redundancy between each other, and the complementary information is not fully explored. Therefore, we aggregate the frame-level features into clip-level feature in this paper.

Specifically, given a clip, a set of frame-level features $\{{v}^{1},{v}^{2},\cdots,{v}^{N}\}$ are extracted through self-supervised video representation learning, then aggregated into a single clip-level feature ${x}$, which is defined as follows:
\begin{equation}
{x} = f_{C}({v}^{1},{v}^{2},\cdots,{v}^{N})
\label{eqa:sic}
\end{equation}

\begin{figure}[!t]
  \begin{center}\includegraphics[width=0.95\linewidth]{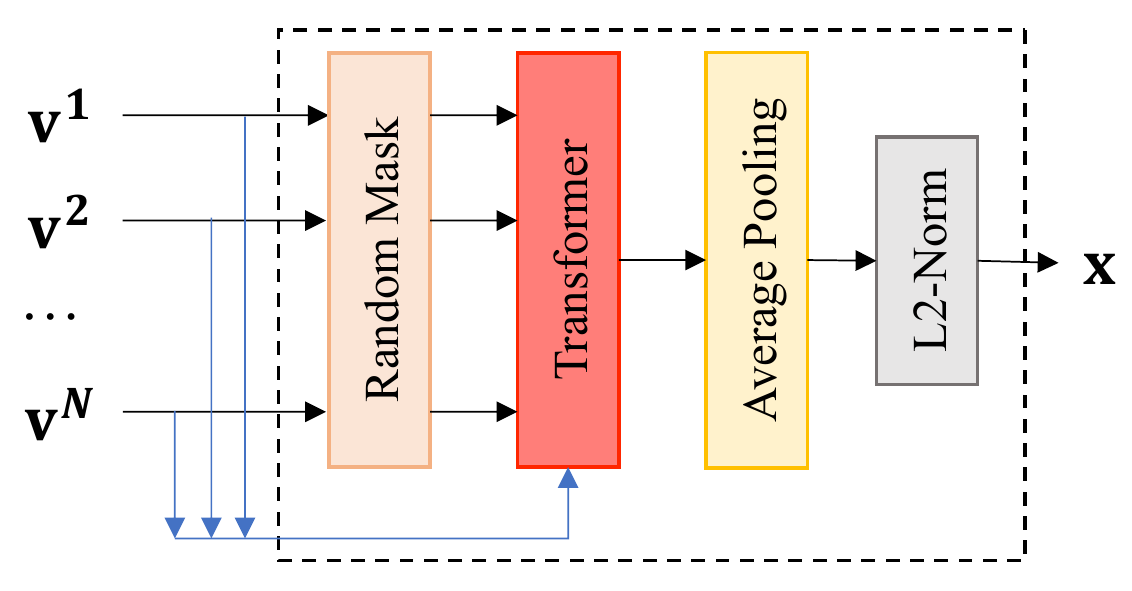}
  \caption{Architecture of our clip-level set transformer network.}
  \label{fig:transformer}
  \end{center}
\end{figure}

To encode the clip-level feature, we propose an adapted Transformer \cite{vaswani2017attention}, called clip-level set transformer network, whose architecture is shown in Figure \ref{fig:transformer}. Instead of directly using Transformer to encode the clip-level feature, we apply the idea of set retrieval \cite{zhong2018compact} in the clip-level encoding. It is noted that we only use  one encoder layer with 8 attention heads, without position embedding. 
It enables our SVRTN approach has the abilities:
(1) More robust. Increase the robustness of the learned clip-level features with the ability of frame permutation and missing invariant.
(2) More flexible. Support more retrieval manners, including clip-to-clip retrieval and frame-to-clip retrieval.

\subsubsection{Clip-level Encoding}
Similar with frame-level encoding in Section \ref{sub:videorepresentationlearning}, given a set of clip-level features $S_C=\{({x}^{b},{x}_{+}^{b})\}_{b=1}^{B}$, where $B$ is the number of clips in a batch, a clip-level constrastive learning is adopted. The loss function is defined as follows:
\begin{equation}
\begin{aligned}
L_{C}(x, x_+) = \frac{1}{B} \sum_{b=1}^{B} &-\mathbb{E}_{P_{d}} \log P(D=1|{x}^{b}, {x}_{+}^{b}) \\
        &-(1-\mathbb{E}_{P_{d}}) \log(1-P(D=1|{x}^{b}, {x}_{+}^{b}))
\label{eqa:sic_loss}
\end{aligned}
\end{equation}
where $P_{d}$ denotes the actual data distribution and $\mathbb{E}_{P_{d}}$ is set to 1 indicates the anchor clip and positive clip share absolutely identical visual semantic.  $P(D=1|{x^b}, {x^b}_{+})$ denotes the posterior probability that ${x}$ with ${x}_{+}$ is from the actual data distribution, its definition is similar with Equation (\ref{eq:pd}).
Clip-level encoding can learn the complementary information from the frames of the video clip via self-attention mechanism of Transformer, and hence the discimination of features via attentively seeing the frames.

\subsubsection{Clip-level Encoding with Masked Frame Modeling}
To increase the robustness of the learned clip-level features, we treat the frames of one clip as a set, and randomly mask some frames in clip-level encoding. For a given clip $C$, we randomly drop some frames to generate a new clip $C'$. Its goal is to eliminate the influence of frame blur or clip cut, and drive the model to have the ability that use any combination of any frames in the clip can retrieval its corresponding clips.

Specially, given a clip-level feature $x$, its new feature after conducting masked frame modeling is denoted as $x'$. Similarly, the corresponding positive clip-level feature and its new feature are denoted as $x_+$ and $x'_+$. Therefore, we need to learn from the following loss functions: $L_C(x, x'_+)$ and $L_C(x', x_+)$. So the final loss function of clip-level set transformer network is defined as follows:
\begin{equation}
    L_{C} = L_{C}(x, x_+) + L_C(x, x'_+) + L_C(x', x_+)
\end{equation}

\subsection{Video Similarity Calculation}
For each video, we first conduct shot boundary detection to segment the videos into shots, and then divide the shots into clips at a fixed time interval, i.e $N$ seconds. Second, the sequence of successive frames is passed through the clip-level set transformer network to generate the clip-level feature. Finally, the clip-level feature is binarized by IsoHash\cite{kong2012isotropic} to further reduce the storage cost and search cost. When retrieving, we measure the clip-to-clip similarities with hamming distance. Given an $M \times N$ clip-to-clip similarity matrix, the video similarity score can be calculated as follows:
\begin{equation}
Sim = \frac{1}{M}\sum\limits_{i=1}^{M}\max\limits_{j\in [1, N]} CS(i, j)
\label{eqa:sim}
\end{equation}
where $CS(i, j)$ denotes the similarity score between clip $i$ and $j$,
and it is calculated as follows:
\begin{equation}
CS(i, j) = \max\limits_{k \in K} \mathcal{H}(i, k) - \mathcal{H}(i, j)
\end{equation}
in which $K$ indicates the entire clip set and $\mathcal{H}(\cdot, \cdot)$ indicates the hamming distance calculation.
\section{Experiments}
\subsection{Datasets}
Our SVRTN approach is trained on our constructed Self-Transformation dataset, and performs evaluations on two challenging video retrieval datasets, namely FIVR-200K and SVD, which focus on fine-grained incident video retrieval and near-duplicate video retrieval respectively. The detailed information is introduced as follows:
\begin{itemize}
\item
\textbf{Self-Transformation} is constructed by collecting videos from video website \footnote{https://www.youku.com/}. It consists of 3,000 hours' videos, and temporal and spatial transformations are performed at training stage.
\item
\textbf{FIVR-200K} \cite{kordopatis2019fivr} consists of 225,960 videos and 100 queries. It is constructed for fine-grained incident video retrieval, including three retrieval tasks: (1) Duplicate scene video retrieval (DSVR) is to retrieval the videos sharing at least one scene that captured by the same camera, regardless of any transformation. (2) Complementary scene video retrieval (CSVR) is to retrieval the videos containing part of the same spatio-temporal segment with different views. (3) Incident scene video retrieval (ISVR) is to retrieval the videos capturing the same event without the same overlapped saptio-temporal segment. We evaluate our SVRTN approach on all the three tasks to verify its effectiveness.
\item
\textbf{SVD} \cite{jiang2019svd} is constructed for short video retrieval task. It consists of 562,013 short videos with the duration less than 60 seconds. It contains 1,206 query videos, and over 30,000 labelled videos in which the negatives have extremely similar but different appearance. Besides, there are more than 500,000 hard negative unlabelled distraction videos to increase the retrieval difficulty.
\end{itemize}

\subsection{Evaluation Metric}
Following \cite{jiang2019svd,kordopatis2019fivr}, we apply the mean average precision (mAP) score to evaluate the video retrieval performance. We first calculate average precision (AP) score for each query, and then calculate their mean value as mAP score.

\subsection{Effectiveness of Reducing Storage and Search Cost}
To verify the effectiveness of our proposed SVRTN approach on reducing the storage and search cost, we compare the storage spaces and search complexities between frame-level retrieval and clip-level retrieval on SVD dataset.
As shown in Table \ref{tab:cost}, the storage of the frame-level features cost 1720.32 MB, while clip-level features only cost 366.98 MB, reducing the storage cost by 78.7\%. It is mainly because that our SVRTN approach first segments the videos into shots, and then divides the shots into clips, finally encodes the clips to represent the videos.

Suppose that the SVD dataset has $m$ queries and $n$ videos, and they are encoded by $M$ frame-level features and $N$ frame-level features respectively. So it needs $\mathcal{O}(M\times N)$ similarity computation.
However, depend on the above analyses, they can be encoded by $\sim\frac{5}M$ and  $\sim\frac{5}N$ clip-level features respectively, so only $\mathcal{O}(\sim\frac{M}{5}\times \sim\frac{N}{5})=\sim\frac{1}{25}\mathcal{O}(M\times N)$ similarity computation is needed. In other words, our SVRTN approach increases the retrieval speed by $\sim25$ times, which verifies that clip-level video retrieval is an efficient retrieval paradigm to reduce the storage cost and search cost. 

\begin{table}[h]
\centering
  \begin{tabular}{|p{1.7cm}<{\centering}|p{2.2cm}<{\centering}|p{2.7cm}<{\centering}|}
    \hline
    Feature &Storage Space& Search Complexity \\
   \hline
   Frame-level & 1720.32 MB & $\mathcal{O}(M\times N)$\\
    \textbf{Clip-level} & \textbf{366.98 MB} & $\bm{\sim\frac{1}{25}\mathcal{O}(M\times N)}$ \\
    \hline 
  \end{tabular}
  \caption{Reduction of storage and search cost on SVD dataset.}
  \label{tab:cost}
\end{table}

\subsection{Comparisons with State-of-the-art Methods}
In this subsection, experimental results and analyses
of comparing our proposed SVRTN approach with the state-of-the-art methods on FIVR-200K and SVD datasets are presented, which are shown in Table \ref{tab:fivr} and Table \ref{tab:svd}. It is noted that we evaluate our SVRTN approach on all the three tasks of FIVR-200K dataset, including DSVR, CSVR, ISVR.

\begin{table*}[h]
\centering
  \begin{tabular}{|p{1.6cm}<{\centering}|p{1.5cm}<{\centering}|p{3cm}<{\centering}|p{1cm}<{\centering}|p{1cm}<{\centering}|p{1cm}<{\centering}|}
    \hline
    Feature &Methods& Feature Dim/\#bits &DSVR & CSVR & ISVR  \\
    \hline
    \multirow{3}{*}{Video-level} & HC\cite{song2013effective} & - & 0.265 & 0.247 & 0.193  \\
    & DML\cite{kordopatis2017near} &500D& 0.398 & 0.378  & 0.309\\
    & TCA\cite{shao2020context} &1024D& 0.570 & 0.553 & 0.473  \\
    \hline
    \multirow{4}{*}{Frame-level} & CNN-L\cite{kordopatis2017near2}& 4096D & 0.710 & 0.675 & 0.572  \\
    &PPT\cite{chou2015pattern}& 4096D & 0.775 & 0.740 & 0.632 \\
    &TN\cite{tan2009scalable} & - & 0.724 & 0.699 & 0.589  \\
    &VisiL\cite{kordopatis2019visil} &9x3840D & 0.892 & 0.841 & 0.702  \\
    & \textbf{SVRTN$_f$}& \textbf{512 bits} & \textbf{0.900} & \textbf{0.858} & \textbf{0.709}  \\
    \hline
    Clip-level &\textbf{SVRTN}& \textbf{512 bits} & \textbf{0.876} & \textbf{0.835} & \textbf{0.686}  \\
    \hline 
  \end{tabular}
  \caption{Comparisons with state-of-the-art methods on all three tasks of FIVR-200K dataset.}
  \label{tab:fivr}
\end{table*}
\subsubsection{Comparisons with Frame-level Retrieval Methods}
We then compare our SVRTN approach with 5 frame-level retrieval methods, which are briefly introduced as follows:
\begin{itemize}
    \item
    CNN-L and CNN-V \cite{kordopatis2017near2} are proposed to convert multiple intermediate CNN features into one vector via layer and vector aggregation schemes respectively.
    \item
    PPT \cite{shao2020context} is a spatio-temporal pattern-based method under the hierarchical filter-and-refine framework.
    \item
    Temporal Network (TN) \cite{tan2009scalable} is proposed to detect the longest shared path between two videos.
    \item
    VisiL \cite{kordopatis2019visil} is proposed to calculate video-to-video similarity from refined frame-to-frame similarity matrices.
\end{itemize}

Compare with frame-level retrieval approach, our SVRTN approach outperforms all state-of-the-art methods except VisiL. It is noted that VisiL adopts a region-aligned matching scheme, which is impractical for large-scale retrieval task due to its low efficiency. While our SVRTN approach still achieves comparable retrieval performance with VisiL under the situation that only using binary codes and no any re-ranking process. Furthermore, when encoding the videos in frame-level features, our SVRTN$_f$ approach can achieve better retrieval performance than VisiL without any complex calculation. 
It is mainly because self-supervised video representation learning can boost the discrimination of the features due to the self-generation of training data, which has strong power in representation learning.
Importantly, no annotated video pairs are needed, which efficiently reduces the expensive cost of manual annotation.

\begin{table}[!t]
\centering
  \begin{tabular}{|p{1.7cm}<{\centering}|p{1.6cm}<{\centering}|p{1.3cm}<{\centering}|p{2cm}<{\centering}|}
    \hline
    Feature &Methods & Feature Dim/\#bits & Top-100 mAP \\
    \hline
    Video-level & DML\cite{kordopatis2017near} & 500D & 0.813 \\
    \hline
    \multirow{3}{*}{Frame-level} &CNN-L\cite{kordopatis2017near2} & 4096D & 0.610 \\
    &CNN-V\cite{kordopatis2017near2} & 4096D & 0.251  \\
    &\textbf{SVRTN$_f$} & \textbf{512 bits} & \textbf{0.871} \\
    \hline
    Clip-level & \textbf{SVRTN} & \textbf{512 bits} & \textbf{0.860} \\
    \hline 
  \end{tabular}
  \caption{Comparisons with state-of-the-art methods on SVD dataset.}
  \label{tab:svd}
\end{table}

\subsubsection{Comparisons with Video-level Retrieval Methods}
We first compare our SVRTN approach with 3 video-level retrieval methods, which are briefly introduced as follows:
\begin{itemize}
    \item 
    Hashing Codes (HC) \cite{song2013effective} is proposed to learn a group of hash functions based on frame-level features, and then combine the hash codes into a single video vector.
    \item
    Deep Metric Learning (DML) \cite{kordopatis2017near} is proposed to early or late fuse the frame-level features into a single video vector, which is then fine-tuned by deep metric learning.
    \item
    Temporal Context Aggregation (TCA) \cite{shao2020context} is proposed to learn a single video vector by aggregating frame-level features with self-attention.
\end{itemize}

Compare with video-level retrieval methods, clip-level retrieval methods needs more storage and search cost. To reduce these costs, we utilize hash codes and measure hamming distances while other methods use floats and measure Euclidean or Cosine distances.
Even so, our SVRTN approach achieves significant improvements by 30.6\%, 28.2\%, 21.3\% mAPs on the DSVR, CSVR and ISVR tasks of FIVR-200K dataset, as well as 4.7\% mAP on SVD dataset, which are shown in Table \ref{tab:fivr} and Table \ref{tab:svd}. 
It are mainly because:
(1) Clip-level feature encoding can extract more abundant and complementary information from the interactions among clip frames.
(2) Clip-level set transformer network can aggregate the frame features in one clip considering their different roles, which takes full advantage of each frame's discrimination, and eliminates the redundancy between the adjacent frames. Besides, it acquires the frame permutation and missing invariant ability with masked frame modeling.

\begin{table}[!ht]
\centering
  \begin{tabular}{|c|c|c|c|c|c|c|}
  \hline
  \multirow{2}{*}{Methods} & \multicolumn{3}{c|}{Transformations} & \multirow{2}{*}{DSVR} & \multirow{2}{*}{CSVR} & \multirow{2}{*}{ISVR}  \\
  \cline{2-4}
  & PT & GT & ET & & & \\
  \hline
  \textbf{SVRTN$_f$} & $\checkmark$ & $\checkmark$ & $\checkmark$ & \textbf{0.900} & \textbf{0.858} & \textbf{0.709} \\
  \hline
  A & $\checkmark$ & $\checkmark$ & & 0.868 & 0.818 & 0.673 \\
  \hline
  B & $\checkmark$ &  & $\checkmark$ & 0.881 & 0.825 & 0.662 \\
  \hline
  C &  & $\checkmark$ & $\checkmark$ & 0.868 & 0.815 & 0.649 \\
  \hline
  \end{tabular}
  \caption{Impacts of different transformations on FIVR-200K dataset.}
  \label{tab:ab_transformation}
\end{table}
\subsection{Ablation Study}

\subsubsection{Effectiveness of Self-supervised Video Representation Learning}
We directly utilize the frame-level feature captured from self-supervised video representation learning to perform video retrieval, results are shown in Table \ref{tab:fivr} and Table \ref{tab:svd} as ``SVRTN$_f$". It outperforms than state-of-the-art methods on both two datasets, which verifies the effectiveness of the self-supervised video representation learning. Due to self-generation of training data, we can generate the training data as much as we want, which breaks the restriction of the expensive manual annotation cost.

Besides, we further evaluate the impact of each transformation on the retrieval performance of self-supervised video representation learning.
The results of three tasks on FIVR-200K dataset are shown in Table \ref{tab:ab_transformation}, where ``PT", ``GT" and ``ET" denote photometric transformation, geometirc transformation and editing transformation respectively, as well as the experiments of ``A", ``B" and ``C" denote training without editing transformation, geometric transformation and photometric transformation respectively. We can observe that ``SVRTN$_f$" with all the three types of transformations achieves the best performance, which verifies that each transformation plays an irreplaceable role on video representation learning.
They provide rich supervision information to drive the model to approximate the real data distribution, which make the learned representation spatial-temporal invariant.

\begin{table}[!t]
\centering
  \begin{tabular}{|p{2cm}<{\centering}|p{1.3cm}<{\centering}|c|c|c|}
  \hline
  \multirow {2}{*}{Methods} &\multirow {2}{*}{SVD} & \multicolumn{3}{c|}{FIVR-200K} \\
   \cline{3-5}
    & & DSVR & CSVR & ISVR \\
  \hline
  CE & 0.854 &0.870 &0.834 &\textbf{0.687} \\
  \hline
  \textbf{CE w/ MFM} & \textbf{0.860} &\textbf{0.876} &\textbf{0.835} &0.686\\
  \hline
  \end{tabular}
  \caption{Effectiveness of clip-level encoding with masked frame modeling.}
  \label{tab:ab_mask}
\end{table}

\begin{table}[!ht]
\centering
  \begin{tabular}{|p{2cm}<{\centering}|p{1.3cm}<{\centering}|c|c|c|}
  \hline
  \multirow {2}{*}{Clip Length} &\multirow {2}{*}{SVD} & \multicolumn{3}{c|}{FIVR-200K} \\
   \cline{3-5}
    & & DSVR & CSVR & ISVR \\
  \hline
  4s & 0.861 & 0.883 & 0.841 & 0.693 \\
  \hline
  6s & 0.867 & 0.881 & 0.837 & 0.688\\
  \hline
  8s & 0.860 & 0.876 &0.835 &0.686\\
  \hline
  \end{tabular}
  \caption{Impact of clip length on clip-level encoding.}
  \label{tab:ab_cliplen}
\end{table}
\subsubsection{Effectiveness of Clip-level Set Transformer Network}
We first evaluate the effectiveness of clip-level encoding with masked frame modeling on FIVR-200K and SVD datasets. Results are shown in Table \ref{tab:ab_mask}, where ``CE" and ``MFM" denote clip-level encoding and masked frame modeling respectively. Clip-level encoding with masked frame modeling coerces the Transformer to learn the correlations between the anchor clip with missing information and positive clip, as well as the anchor clip and positive clip with missing information, which makes the transformer more robust and not sensitive to the frame missing. So it improves the discrimination and robustness of the learned clip-level feature, and achieves better performance. 

Besides, we evaluate the impact of clip length to the retrieval performance of clip-level set transformer network. Table \ref{tab:ab_cliplen} shows the results of different clip length settings on FIVR-200K and SVD datasets. We can observe that our clip-level set transformer network is not very sensitive to the clip lengths. It is mainly because that we apply masked frame modeling in clip-level encoding, which drives the model to have the ability that any combination of any frames in the clip can retrieval its corresponding clips. So to balance the retrieval accuracy and efficiency, we set the clip length as 8s in our experiments.

\subsection{Exploration of Flexible Retrieval Manners}
As mentioned above, clip-level set transformer network provides more flexible retrieval manners, i.e. clip-to-clip retrieval and frame-to-clip retrieval. So we explore their retrieval performance on SVD dataset, as shown in Table \ref{tab:ab_flexible}. The videos in database are all encoded in clip-level features, only different in query encoding. We can see that use more fine-grained features (i.e. frame-level) can achieve better retrieval performance, which further verifies the effectiveness of clip-level encoding with masked frame modeling, which can driven our SVRTN approach learn both clip-level and frame-level features. With more flexible retrieval manners, our SVRTN approach has more application prospects.
\begin{table}[ht]
\centering
  \begin{tabular}{|p{2.2cm}<{\centering}|p{2.2cm}<{\centering}|p{2.4cm}<{\centering}|}
  \hline
  Query&Database & Top-100 mAP \\
  \hline
  Clip & Clip &0.860  \\
  \hline
  Frame & Clip & 0.871\\
  \hline
  \end{tabular}
  \caption{Results of different retrieval manners.}
  \label{tab:ab_flexible}
\end{table}

\section{Conclusion}
This paper proposes the SVRTN approach to encode the video in clip-level representation with self-supervised learning to reduce the expensive cost of manual annotation, storage space and similarity search . It consists of two components: 
(1) Self-supervised video representation learning is proposed to automatically generate the pairs of the videos and their transformations as supervision information, which reduces the heavy labor consumption in annotating. Besides, with more self-generated supervised data, the discrimination and robustness of the learned feature are increased.
(2) Clip-level set transformer network is proposed to reduce the redundancy of the frames in a clip, as well as learn the complementary and discriminative information from the interactions among clip frames. Besides, clip-level encoding with masked frame modeling make the model frame permutation and missing invariant, and support more flexible retrieval manners.
Comprehensive experiments on two challenging video retrieval datasets, namely SVD and FIVR-200K, verify the effectiveness of our SVRTN approach, which achieves the best performance of video retrieval on accuracy and efficiency.

The future works will lie in two aspects: (1) How to design more efficient self-supervised learning? (2) How to transfer the motion information to RGB frames in training, but only use RGB frames in retrieval. Both of them will be explored to further improve the video retrieval performance.

\balance
{\small
\bibliographystyle{ieee_fullname}
\bibliography{egbib}
}

\end{document}